\documentclass[10pt,twocolumn,letterpaper]{article}

\usepackage[pagenumbers]{cvpr} 

\usepackage{fixmetodonotes}

\newif\ifshowedits

\newcommand{\addeditor}[3]{%
  \definecolor{#1color}{rgb}{#3}
  \expandafter\newcommand\csname #1\endcsname[1]{%
  \ifshowedits
    {\color{#1color} ##1}%
  \else
    {##1}%
  \fi
  }%
  \expandafter\newcommand\csname #1rmk\endcsname[1]{%
  \ifshowedits
    {\color{#1color} {\bf [#2: ##1]}}
  \fi
  }%
  \expandafter\newcommand\csname #1rpl\endcsname[2]{%
  \ifshowedits
    {\color{#1color} ##1 \sout{##2}}
  \else
    {##1}
  \fi
  }%
}

\addeditor{eloi}{EZ}{0.2, 0.2, 0.75}
\addeditor{valentin}{VG}{0.1, 0.55, .2}
\addeditor{ahmad}{AR}{0.75, 0.2, 0.2}

\showeditstrue

\usepackage{mathtools}
\usepackage{multirow}

\definecolor{cvprblue}{rgb}{0.21,0.49,0.74}
\usepackage[pagebackref,breaklinks,colorlinks,allcolors=cvprblue]{hyperref}

\def\paperID{7427} 
\def\confName{CVPR}
\def\confYear{2026}

\title{MAD: Motion Appearance Decoupling for efficient Driving World Models}

\author{Ahmad Rahimi $^{*\text{ }1}$ \hspace{0.3cm}
Valentin Gerard $^{*\text{ }1}$ \hspace{0.3cm}
\'Eloi Zablocki $^{2}$ \hspace{0.3cm} 
Matthieu Cord $^{2,\text{ }3}$  \hspace{0.3cm} 
Alexandre Alahi $^1$\\[0.1cm]
$^1$ Ecole Polythechnique Federal de Lausanne \quad $^2$ valeo.ai \quad $^3$ Sorbonne Université\\
{\tt\small first.last@epfl.ch} \quad {\tt\small first.last@valeo.com} \quad {\tt\small matthieu.cord@sorbonne-universite.fr} \\
}

\begin{document}
\maketitle

\begin{abstract}

Recent video diffusion models generate photorealistic, temporally coherent videos, yet they fall short as reliable world models for autonomous driving, where structured motion and physically consistent interactions are essential. Adapting these generalist video models to driving domains has shown promise but typically requires massive domain-specific data and costly fine-tuning.
We propose an efficient adaptation framework that converts generalist video diffusion models into controllable driving world models with minimal supervision. The key idea is to decouple \emph{motion learning} from \emph{appearance synthesis}. First, the model is adapted to predict structured motion in a simplified form: videos of skeletonized agents and scene elements, focusing learning on physical and social plausibility. Then, the same backbone is reused to synthesize realistic RGB videos conditioned on these motion sequences, effectively “dressing” the motion with texture and lighting.
This two-stage process mirrors a reasoning-rendering paradigm: first infer dynamics, then render appearance. Our experiments show this decoupled approach is exceptionally efficient: adapting SVD, we match prior SOTA models with \textbf{less than 6\% of their compute}. Scaling to LTX, our MAD-LTX model outperforms all open-source competitors, and supports a comprehensive suite of text, ego, and object controls. Project page: \url{https://vita-epfl.github.io/MAD-World-Model/}

\end{abstract}

\section{Introduction}
\label{sec:intro}

\begin{figure}[!t]
  \centering
  \includegraphics[width=1\columnwidth, trim=2.3cm 0cm 0.5cm 0, clip]{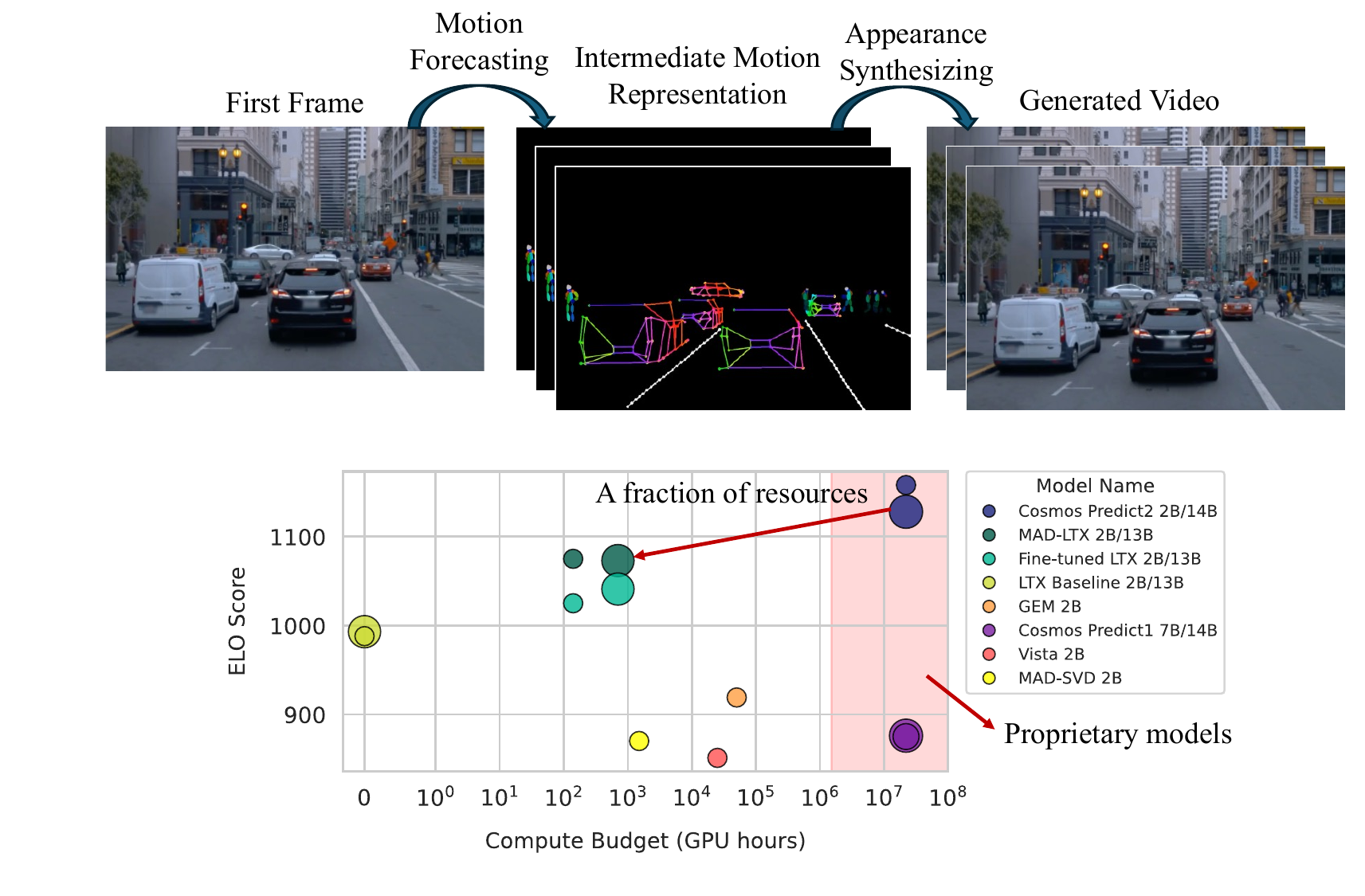}
  \caption{(Top) Our decoupled two-stage pipeline: a Motion Forecaster first generates an abstract intermediate pose representation, which is then used by an Appearance Synthesizer to render the final, photorealistic video. (Bottom) Our method (MAD-LTX) achieves a state-of-the-art quality while requiring a fraction of the training compute budget compared to prior SOTA driving models.} 
  \label{fig:teaser}
\end{figure}

Professional animators do not begin by drawing final, photorealistic frames. They first create an \textit{animatic}, a sequence of simple, timed-out sketches, to perfect the core of the story: its pacing, composition, and, most critically, its \textit{interactions} and \textit{motion}~\citep{glebas2012directing}. The final, high-fidelity rendering of light, shadow, and texture is the last, almost mechanical, step. This principle of decoupling a complex creative task into \textit{dynamics} and \textit{rendering} is a powerful and fundamental approach.

We apply this \textit{animator} principle to the central challenge of creating autonomous driving (AD) world models.
Unlike general-purpose video generation models (VGMs), which are trained on large, diverse internet datasets to generate generic videos and primarily excel at visual realism, driving models must \emph{also} master the physically grounded, multi-agent \emph{dynamics} of traffic; a task that VGMs fail to capture off the shelf.
This reveals the core difficulty: a driving world model must simultaneously master two coupled, complex responsibilities: photorealistic appearance and physical/social dynamics.

Adapting existing general VGMs to the driving domain is therefore prohibitively expensive. Recent efforts illustrate this: VISTA~\citep{gao2024vista} and GEM~\citep{hassan2025gem} consume 25,000 and 50,000 GPU-hours to fine-tune the SVD~\citep{blattman2023svd} backbone. More powerful models like Cosmos-Predict~\citep{nvidia2025cosmosworldfoundationmodel} are trained from scratch on massive, proprietary datasets, requiring compute resources that place them beyond the reach of most research labs. This high cost of adaptation prevents the AD community from fully leveraging the rapid, ongoing advances in general video modeling.

We therefore propose \textbf{MAD} (\textbf{M}otion-\textbf{A}ppearance \textbf{D}ecoupling), a new methodology for efficiently adapting any general VGM into a controllable driving world model. As illustrated in \autoref{fig:teaser}, we adapt a single VGM backbone to specialize in the two sequential stages of the animator's workflow, using two lightweight LoRA~\citep{hu2022lora} adapters:
\begin{enumerate}
    \item A \textbf{Motion Forecaster} masters the \textit{animatic} stage, learning to generate an abstract, skeletonized video that captures only the scene's dynamics. This representation does not require new manual labels; we generate it at scale by applying off-the-shelf pose extractors~\citep{kreiss2022openpifpaf, yang2023dwpose} as pseudo-labelers to our video data.
    \item An \textbf{Appearance Synthesizer} masters the \textit{rendering} stage, learning to “dress” this motion by rendering the final photorealistic video conditioned on the abstract skeleton poses.
\end{enumerate}

By having a \textit{single model} perform both sequential tasks, our framework is conceptually similar to the 'chain-of-thought' paradigm, where a model "reasons" by generating intermediate steps (the motion "animatic") before producing the final "answer" (the rendered video). This single-model, two-step approach is further enhanced by a second key principle: maximal reuse of the base model's learned representations. Instead of engineering new, complex conditioning mechanisms, we project all control signals, including the intermediate poses, the first frame, and even our novel ego-motion representation, into the model's native visual latent space using its own pre-trained VAE. This allows the model to learn new control tasks while `speaking' a visual language it already understands.

We validate this framework by creating MAD-SVD and MAD-LTX. As a proof-of-concept, MAD-SVD builds on the SVD backbone and achieves quality competitive with VISTA \citep{gao2024vista} and GEM \citep{hassan2025gem} while using \textbf{less than 6\% of their compute and data}. We then scale this methodology to create \textbf{MAD-LTX}, an open-source, state-of-the-art driving model building on the LTX backbone. MAD-LTX outperforms all previous open driving models and achieves generation quality comparable to the top-performing, open-weight Cosmos Predict 2 \citep{nvidia2025cosmosworldfoundationmodel}. This is accomplished while using only \textbf{a fraction of compute and data budget} of competitors, and supporting a combination of text, ego-motion, and object-motion controls.

\paragraph{Contributions.}
Our main contributions are:
\begin{itemize}
    \item A novel methodology that decouples motion forecasting from appearance synthesis, enabling highly efficient adaptation of general VGMs.
    \item An efficient adaptation framework that reuses a single backbone with lightweight LoRAs and leverages the model's native VAE for all conditioning, proving more efficient than standard fine-tuning.
    \item \textbf{MAD-LTX}, a new open-source, SOTA driving world model (2B \& 13B) that is the fastest open-source driving world model and supports a comprehensive suite of text, ego-motion, and object-motion controls.
\end{itemize}

\section{Related Work}
\label{sec:related}
\subsection{Driving World Models}

World models \cite{https://doi.org/10.5281/zenodo.1207631} aim to learn action-conditioned dynamics of an environment, predicting plausible futures while capturing underlying physical regularities and agent behaviors that generalize to rare or unseen situations. Beyond future prediction, such models support a wide range of downstream applications, including policy learning \cite{hafner2020dreamcontrollearningbehaviors} where agent can plan inside the learned model rather than in the real environment, representation learning \cite{bartoccioni2025vavimvavamautonomousdriving, bardes2024revisitingfeaturepredictionlearning} via predictive objectives, and large-scale synthetic data generation \cite{ren2025cosmosdrivedreamsscalablesyntheticdriving}.

In autonomous driving, world models have been instantiated using diverse input parameterizations that reflect different modeling objectives. Semantic BEV and occupancy-centric \cite{li2025unisceneunifiedoccupancycentricdriving, zuo2024gaussianworldgaussianworldmodel, chen2025occprophetpushingefficiencyfrontier} approaches forecast the evolution of road topology, traffic participants, and freespace in a structured mid-level representation aligned with classical planning pipelines. In contrast, LiDAR-based \cite{liang2025learninggenerate4dlidar, bian2025dynamiccitylargescale4doccupancy} and implicit 4D occupancy-flow models operate directly on dense geometric structure to enable high-fidelity reconstruction. Moving beyond structured ontology-based scene abstractions, video-based driving world models directly generate future ego-centric RGB observations.

These approaches differ in their sensor assumptions: multi-view methods \cite{hu2023gaia1generativeworldmodel, russell2025gaia2controllablemultiviewgenerative, Yang_2025_ICCV}  exploit surround-camera geometry for 3D-consistent scene generation, whereas single-view models \cite{hassan2025gem, gao2024vista, nvidia2025cosmosworldfoundationmodel,zhang2025eponaautoregressivediffusionworld, yang2025resimreliableworldsimulation,bartoccioni2025vavimvavamautonomousdriving} operate purely from monocular ego video, offering a critical scalability advantage, as monocular driving video can be collected at scale, even from unconstrained Internet sources \cite{yang2024opendv, alam2024crossing}. Our work focus on this single-view paradigm, which naturally connects with generalist video generation models literature.

\subsection{Generalist Video Generation Models}

Generalist video generation models \cite{wan2025wanopenadvancedlargescale, kong2025hunyuanvideosystematicframeworklarge, yang2025cogvideoxtexttovideodiffusionmodels, hacohen2025ltx, blattman2023svd} are commonly pretrained on large, weakly curated Internet-scale video corpora, which endow them with broad visual and temporal priors. 

Despite their impressive visual realism, recent benchmarks \cite{motamed2025generativevideomodelsunderstand, wang2025videoversefart2vgenerator, qin2024worldsimbenchvideogenerationmodels, zhang2025morpheusbenchmarkingphysicalreasoning} show that such models struggle with physical consistency, multi-object dynamics, and causal interactions. They tend to reproduce statistically dominant motion patterns rather than adapting to perturbed environments, causing systematic failures in scenarios requiring agent coordination or physics reasoning. These limitations highlight a gap between generalist video generators and the physically grounded world-modeling capabilities needed for reliable driving simulation.

To compensate for these limitations, several domain adaptation strategies have emerged, each targeting the lack of diverse and physically grounded motion patterns in generalist models. Models trained on weakly curated crowd-sourced driving videos \citep{yang2024opendv}, such as VISTA \cite{gao2024vista} and GEM \cite{hassan2025gem}, adapt SVD \cite{blattman2023svd} through large-scale fine-tuning on OpenDV \cite{yang2024opendv}, requiring substantial computational budgets (approximately 25k and 50k GPU-hours, respectively). Other approaches incorporate synthetic trajectories to enrich multi-object dynamics.  ReSim \cite{yang2025resimreliableworldsimulation} adapts CogVideoX \cite{yang2025cogvideoxtexttovideodiffusionmodels} via large-scale, 13k GPU-hour fine-tuning, augmenting training dataset with CARLA \cite{dosovitskiy2017carlaopenurbandriving}-generated trajectories. At the industrial scale, models such as GAIA \cite{hu2023gaia1generativeworldmodel, russell2025gaia2controllablemultiviewgenerative} and Cosmos-Predict \cite{nvidia2025cosmosworldfoundationmodel} circumvent the adaptation bottleneck entirely by training driving-oriented world models from scratch on massive proprietary datasets.

Together, these efforts underscore that accurately modeling multi-object dynamics remains highly data- and compute-intensive. Our approach aims to reduce this burden by enabling efficient transfer from generalist video generators to driving world models.

\subsection{Prior works factorizing motion and appearance}

Recent findings show that two-stage video generation pipelines, where an intermediate motion representation is predicted before appearance synthesis, improve temporal coherence, especially in sequences exhibiting large motion.
In these approaches, two diffusion models are applied sequentially: the first one predicts a coarse motion signal (e.g., an object-level mask video \cite{yariv2025throughthemaskmaskbasedmotiontrajectories}, dense pixel-wise motion fields \cite{shi2024motioni2vconsistentcontrollableimagetovideo}, or flow- and depth-based geometric cues \cite{liang2024movideomotionawarevideogeneration}), and the second one synthesizes the final video conditioned on this predicted motion representation. We adopt a similar decoupling of motion prediction and appearance synthesis through pose based representation, and show that it enables efficient adaptation to driving scenes with complex, structured multi-agent motion.

In the driving domain, Epona \cite{zhang2025eponaautoregressivediffusionworld} adopts an autoregressive formulation in which a dedicated trajectory predictor models future ego motion, while a video generator produces the corresponding visual sequence. This coupling stabilizes long-horizon predictions and allows the model to incorporate planning signals. However, because the intermediate motion representation is limited to the ego vehicle, it cannot capture the multi-agent interactions required to fully disentangle motion from appearance in complex driving environments. To the best of our knowledge, we are the first to fully separate multi-agent motion prediction from appearance synthesis in a driving world model.

A different family of methods focus on appearance-level editing for data augmentation, where visual attributes such as weather, illumination, or style are modified while keeping the underlying scene structure fixed. These models \cite{ren2025cosmosdrivedreamsscalablesyntheticdriving, Yang_2025_ICCV, nvidia2025cosmostransfer1conditionalworldgeneration} condition synthesis on HD-maps, semantic layouts or trajectories extracted from real videos, which provide a static description of geometry and motion. Although this connects to the appearance-synthesis stage of our pipeline, their conditioning structure is extracted rather than predicted, thus placing these methods in an editing task rather than a predictive setting.

\section{MAD Method}
\label{sec:method}

\begin{figure*}[t]
  \centering
  \includegraphics[width=\textwidth]{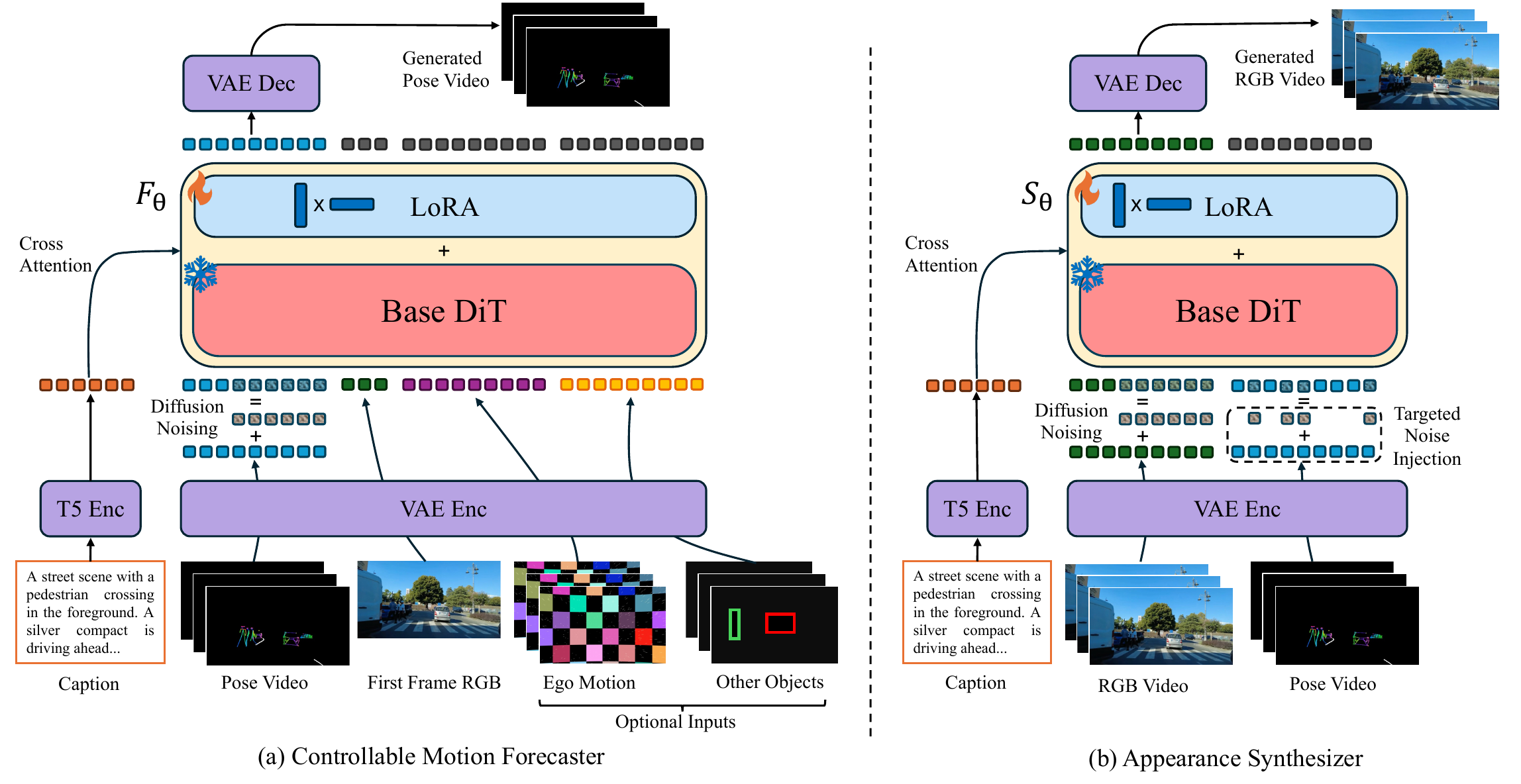}
  \caption{\textbf{Training pipeline of our two models.} a) depicts how the pose video is noised in the diffusion process and given as input to the LoRA of the base DiT model, namely $F_\theta$, together with different conditions such as text captions, first frame in RGB and optionally, our visual representation of the ego motion and other objects location. The denoised pose tokens can be decoded back to the generated pose video at inference time. Similarly, b) shows the training scheme for our appearance synthesizer model $S_\theta$, which takes the same text tokens and the visual tokens of the pose video, noised where the token contains skeletons in the ``targeted noise injection'' block, and denoises RGB video tokens. At inference time, both models are used sequentially, with the generated pose video of the first model being used to control the generation of the second one. }

  \label{fig:method}
\end{figure*}

This section first defines the problem formulation and our core design intuition in \autoref{sec:model:formulation}. It follows by describing our intermediate motion representation for driving scenes (\autoref{sec:model:motion}), then details our methodology for the design and training of the motion forecaster (\autoref{sec:model:forecaster}), its optional control signals (\autoref{sec:model:guidance}), and the appearance synthesizer (\autoref{sec:model:appearance}).

\subsection{Motivation}
\label{sec:model:formulation}

Generative models, such as diffusion or flow-based models $f_\theta$, learn to map a simple prior distribution $\mathcal{N}$ (e.g., a Gaussian) to a complex data distribution $\mathcal{X}$. This mapping can be denoted as $\mathcal{N} \xrightarrow[]{f_\theta} \mathcal{X}$. For video generation, unconditional sampling from the video distribution $\mathcal{X}$ is of limited use. Practical models are conditioned on a set of control signals $\mathbb{C}$, such as the first frame $\mathcal{I}$ or a text prompt $\mathcal{T}$, to guide the synthesis process: $\mathcal{N} \xrightarrow[\mathbb{C}]{f_\theta} \mathcal{X}$.

Driving world models are a specific instance of this setting: they must generate realistic driving scenarios conditioned on high-level ego-vehicle commands. This is particularly challenging as it requires jointly modeling two complex and coupled factors: 1) the dynamic, realistic motion of all actors, and 2) the photorealistic synthesis of pixel-level appearance. Modeling both simultaneously is exceptionally data-intensive and computationally demanding.

Our core intuition is to decouple this joint problem. We first introduce an intermediate motion representation, $\mathcal{M}$, which abstracts away photorealistic appearance to represent only the scene's underlying motion. We then decompose the generation process into two distinct, more manageable stages:
\begin{enumerate}
    \item A \textbf{Motion Forecaster} ($F_\theta$) generates a future motion representation $\mathcal{M}$ from noise, conditioned on motion-centric controls $\mathbb{C}_{\text{motion}}$ (e.g., text, ego-motion, or first RGB frame providing context): $\mathcal{N} \xrightarrow[\mathbb{C}_{\text{motion}}]{F_{\theta}} \mathcal{M}$.
    \item An \textbf{Appearance Synthesizer} ($S_\theta$) then renders the final, photorealistic video $\mathcal{X}$, conditioned on the generated motion $\mathcal{M}$ and appearance-based controls $\mathbb{C}_{\text{appearance}}$ (e.g., the first RGB frame): $\mathcal{N} \xrightarrow[\{\mathcal{M}, \mathbb{C}_{\text{appearance}}\}]{S_{\theta}} \mathcal{X}$.
\end{enumerate}

This two-stage pipeline forms our methodology. A key aspect of our approach is to avoid training $F_\theta$ and $S_\theta$ from scratch. We argue that large-scale, general video generation models already possess significant latent knowledge of both motion dynamics and visual appearance.
We therefore adapt a pre-trained base model to these two specialized tasks by fine-tuning with LoRA~\citep{hu2022lora}. This strategy maximally leverages the base model's powerful visual priors, drastically reducing the training burden and ensuring our conditioning signals operate within a familiar latent space.

\subsection{Intermediate Motion Representation}
\label{sec:model:motion}

Our intermediate motion representation, $\mathcal{M}$, is designed to isolate the motion dynamics of the scene by abstracting away photorealistic textures. We define $\mathcal{M}$ as a ``pose video", a sequence of frames rendered on a black background, depicting the skeletons of dynamic agents (cars, pedestrians) and key static elements (lane lines). Joints and edges are plotted in distinct colors to differentiate agent types. To generate our training data, we extract these poses from real videos using off-the-shelf models~\citep{yang2023dwpose, kreiss2022openpifpaf} to create pseudo-labels.

\begin{figure}[t]
  \centering
  \includegraphics[width=1\columnwidth]{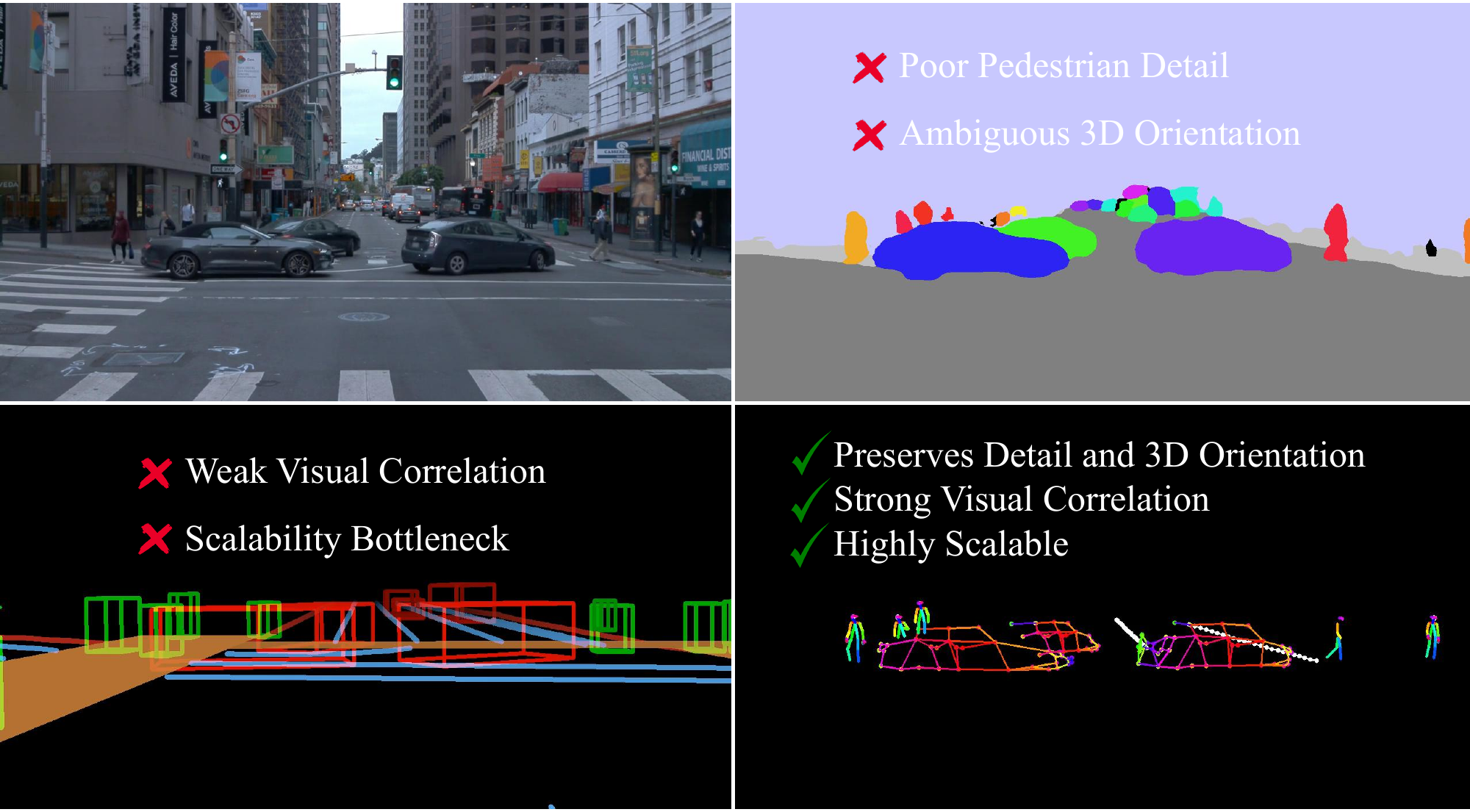}
  \caption{\textbf{Comparison of intermediate representations $\mathcal{M}$.} HDMaps are abstract and non-scalable. Panoptic segmentation is 3D-blind and struggles with pedestrian details. Our pose-based representation provides a scalable, 3D-aware, and object-centric structure that is ideal for both forecasting and synthesis.}
  \label{fig:rep-comparison}
\end{figure}

As shown in \autoref{fig:rep-comparison}, we explored several representations. While HDMaps (3D bounding boxes) are common in classical motion forecasting, they have three drawbacks for our generative setting: 1) They are too abstract, making it difficult for the VGM to realize the bounding box represents a car or pedestrian. Hence the model can not leverage its learned priors, which hurts knowledge transfer. 2) This abstraction gap also burdens the appearance synthesizer, which must learn to correlate simple boxes with complex vehicle appearances. 3) They are not scalable, as they require a full AV perception stack and are limited to small, annotated datasets. In contrast, both panoptic segmentation and our pose representation are highly scalable, as pseudo-labels can be generated for any video. However, as \autoref{fig:rep-comparison} shows, panoptic segmentation, while pixel-accurate, is largely 2D and struggles to capture 3D orientation or detailed pedestrian articulation. Our pose representation strikes the best balance: it is scalable, 3D-aware, and provides an object-centric structure that aligns well with the priors of VGMs, simplifying both forecasting and synthesis.

\subsection{Motion Forecaster}
\label{sec:model:forecaster}
Our Motion Forecaster, $F_\theta$, is designed to generate the future pose video $\mathcal{M}$. While standard motion forecasting often relies on Transformer-based architectures~\citep{xu2024modular4cast,feng2024unitraj}, these approaches present several fundamental limitations for our \emph{generative} world-model task. First, they are typically deterministic, designed to forecast the future states of \emph{existing} agents, and are incapable of ``imagining'' new agents entering the scene. Second, they often require a history of multiple observation states, whereas our goal is to generate from a \emph{single} frame. Finally, such models would need to learn visual context (like the first RGB frame) from scratch, a highly data-intensive process. We therefore bypass these issues by reformulating the task as a \emph{generative} one, fine-tuning a pre-trained VGM on our pose videos $\mathcal{M}$. This generative approach not only learns the distribution of plausible driving motions exceptionally fast but also inherently supports the generation of new agents. It also allows us to leverage the VGM's strong, pre-trained visual priors to understand the first RGB frame, as we detail below.

A key methodological finding is the importance of operating within the VGM's “comfort zone”. While the model can technically handle arbitrary inputs, we found that training at its native resolution and frame rate was critical. Downsampling the data (e.g., to reduce FPS or resolution) significantly degraded forecasting quality, as it forced the model to learn new, out-of-distribution motion priors, which requires substantially more data. By respecting the model's pre-trained priors, we minimize the adaptation cost.

As illustrated in \autoref{fig:method} panel (a), the forecaster $F_\theta$ is trained as a latent diffusion model, conditioned on the initial state of the scene. A core principle of our design is to leverage the base model's pre-trained VAE for all visual conditioning, eliminating the need for new adapter networks. The forecaster is conditioned on the initial scene state and an optional text prompt, which are integrated into the DiT as follows:
\begin{itemize}
    \item \textbf{Noisy Latent \& First Pose Frame:} The pose video $\mathcal{M}$ is encoded into latents $z = E_{\text{VAE}}(\mathcal{M})$. These latents $z_{1...T}$ are noised according to the base model's training schedule. To condition on the initial state, the latent features $z_0$ corresponding to the first pose frame are kept clean (i.e., not noised) and concatenated to the noised sequence.
    
    \item \textbf{Text Prompt:} Unlike prior driving models that omit text \cite{gao2024vista, hassan2025gem}, we retain it for prompt-based motion control. We adopt the base model's standard text conditioning, encoding prompts with a T5~\citep{raffel2020t5} encoder and integrating them via cross-attention.
    
    \item \textbf{First RGB Frame:} While $z_0$ provides the initial pose, the first RGB frame $\mathcal{I}_0$ contains crucial context (e.g., traffic light state, object presence) that may be lost during pose extraction. We encode $\mathcal{I}_0$ using the VAE and concatenate its latent representation $c_{\text{rgb}} = E_{\text{VAE}}(\mathcal{I}_0)$ to the DiT input sequence as additional context. This cross-modal conditioning is remarkably data and compute efficient, as the VGM backbone already possesses a deep understanding of RGB images from its pre-training.
\end{itemize}

\begin{figure}[t]
  \centering
  \includegraphics[width=1\columnwidth]{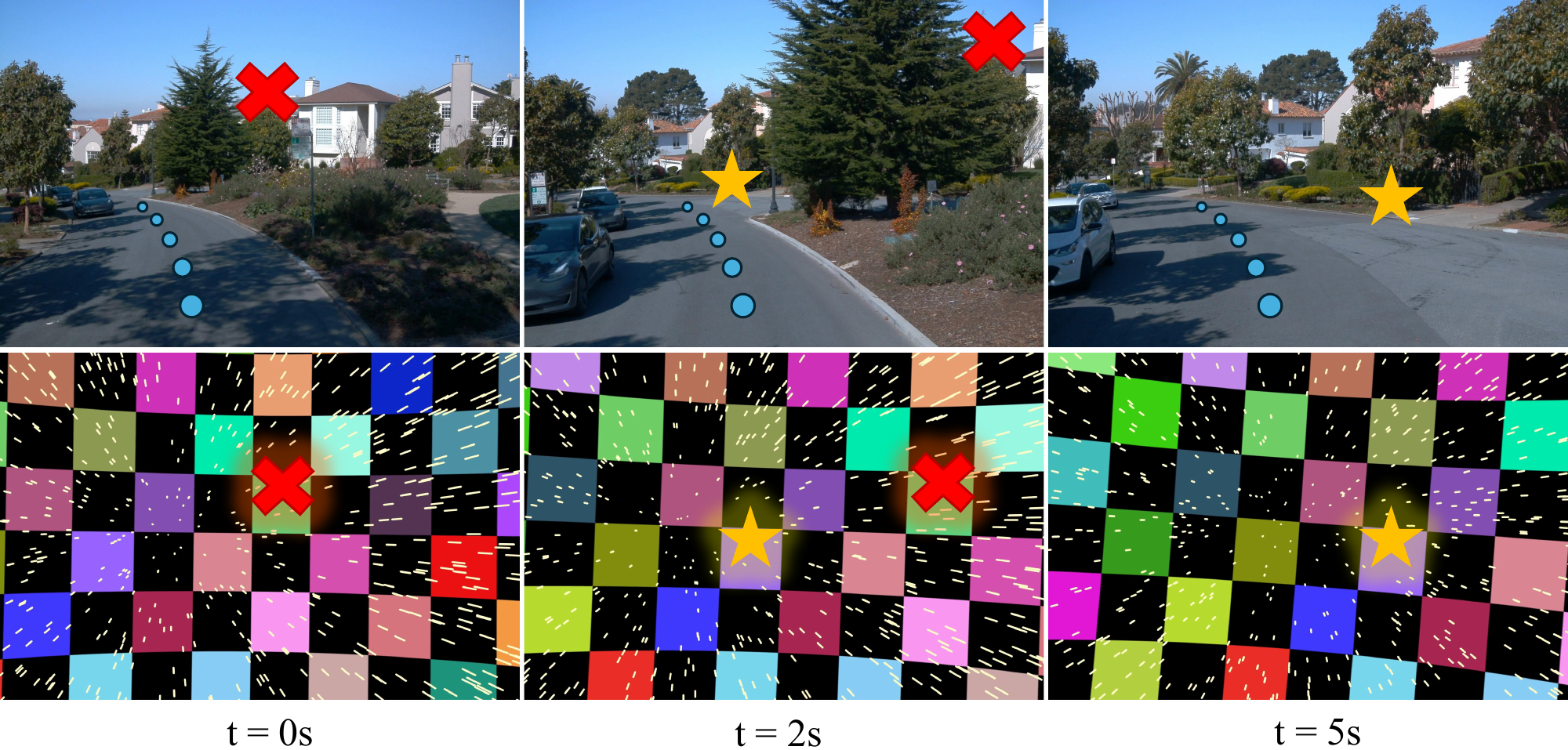}
  \caption{\textbf{Illustration of our visual ego-motion representation}, showing three frames from a car turning left. The ego-camera is surrounded by a static, textured sphere and dust particles. Rotations can be inferred from the background's apparent motion (moving right), while speed is encoded by the parallax motion of the particles. Blue dots represent the car's trajectory, and the markings show how the movement between sphere texture and static elements in the background are aligned.}
  \label{fig:camera-motion-illustration}
\end{figure}

\subsection{Controllable Motion Conditioning}
\label{sec:model:guidance}

A key requirement for a driving world model is controllability. Our framework supports this through the text prompts (as described in \autoref{sec:model:forecaster}) and two additional, \emph{conditioning signals} that are fed to the Motion Forecaster:

\begin{itemize}
    \item \textbf{Ego-Motion Control:} To control the ego-vehicle, we devise a novel visual representation for its 3D camera pose. We render a video $\mathcal{V}_{\text{ego}}$ from the ego-camera's perspective as it moves within a synthetic world. As illustrated in \autoref{fig:camera-motion-illustration}, this world consists of a static sphere with a colored checkerboard texture and static dust particles. Rotations are inferred from the apparent motion of the checkerboard (e.g., the background moves right as the car turns left), while translations are captured by the parallax motion of the dust particles. As highlighted by the markings in \autoref{fig:camera-motion-illustration}, the apparent motion of the background texture is designed to closely correspond to the motion of static objects in the real video. We find this visual representation allows the VGM to easily correlate the control input with the expected evolution of the scene. This video is encoded by the VAE, and its latent $c_{\text{ego}} = E_{\text{VAE}}(\mathcal{V}_{\text{ego}})$ is concatenated to the DiT sequence.
    
    \item \textbf{Object Movement Control:} To enable fine-grained scenario creation, we introduce an optional control for object motion. We first extract 2D bounding boxes from our pose data and run a tracker to get object trajectories. During training, we randomly select up to 5 tracks, render their bounding box trajectories onto a black background to create a sparse control video $\mathcal{V}_{\text{obj}}$, and encode it via the VAE ($c_{\text{obj}} = E_{\text{VAE}}(\mathcal{V}_{\text{obj}})$). This sparsity in training mimics the inference-time use case, where a user typically wants to control only a few specific agents.
\end{itemize}

\subsection{Appearance Synthesizer}
\label{sec:model:appearance}

The second stage of our pipeline is the Appearance Synthesizer, $S_\theta$, which is trained to generate the final photorealistic video $\mathcal{X}$. It is conditioned on the first RGB frame $\mathcal{I}_0$, a text prompt $\mathcal{T}$, and the intermediate motion representation $\mathcal{M}$. The text and first-frame conditioning follow the same method as the forecaster model.

The primary conditioning signal is the pose video $\mathcal{M}$, which is encoded $c_{\text{pose}} = E_{\text{VAE}}(\mathcal{M})$ and concatenated to the DiT input. A key challenge arises from the discrepancy between training and inference: during training, the model is conditioned on clean, ground-truth pose videos $\mathcal{M}_{\text{gt}}$ extracted using pseudo-labelers, but at inference, it receives the predicted $\mathcal{M}_{\text{pred}}$ from the motion forecaster, which contain artifacts (e.g., blurry or bent lines).

To bridge this domain gap, we simulate inference-time imperfections during training, as shown in \autoref{fig:method} (b) (targeted noise injection block). We apply a targeted noising strategy to the ground-truth pose latents $c_{\text{pose}}$. This choice is deliberate: we add noise in the \textit{latent space} because $F_\theta$ is a latent diffusion model, and its artifacts (blurriness, bent lines) are more faithfully represented as latent-space noise rather than simple pixel-space noise. Furthermore, we observe that the background generated by $F_\theta$ is not noisy. We therefore add Gaussian noise with a random variance $\sigma \sim U(0, 0.3)$ only to a sparse set of the latent features corresponding to the rendered skeleton parts and keep the black background latents clean. This targeted strategy forces $S_\theta$ to become robust to imperfections in the motion structure without corrupting the clean background, which we find improves the quality of the final synthesized video.

\section{Experiments}
\label{sec:experiments}

This section validates our proposed methodology. We first provide a detailed overview of our datasets, evaluation protocols, and baseline models in \autoref{sec:exp:setup}. We then present \autoref{sec:exp:SVD-rep}, a proof-of-concept implementation, MAD-SVD, which demonstrates our method's efficiency by adapting a general video model to the driving domain. \autoref{sec:exp:MAD-LTX} details our primary contribution, MAD-LTX, which scales our approach to a more recent foundation model to achieve state-of-the-art generation quality and controllability. Finally, \autoref{sec:exp:ablations} presents a series of ablation studies that justify our key design choices.

\subsection{Experimental Setup}
\label{sec:exp:setup}

\paragraph{Dataset and Preprocessing:}
Our primary dataset, OpenDV \citep{yang2024opendv}, consists of 1700 hours of driving footage from YouTube. We preprocess videos to 24fps at a 1056$\times$704 resolution. We then extract our intermediate motion representation $\mathcal{M}$ by using OpenPifPaf~\cite{kreiss2022openpifpaf} for car and lane poses and DWPose~\cite{yang2023dwpose} for human poses. We segment the videos into 5-second clips (120 frames) with a 3-second overlap. To improve data quality, we filter out 50\% of the clips with the lowest object counts. OpenDV provides distinct training and validation video splits. To create our dataset, we sample 100,000 clips exclusively from the training videos and 5,000 clips exclusively from the validation videos. This ensures no data leakage between our training and validation sets, despite the 3-second overlap in clip extraction. To generate text conditioning $\mathcal{T}$, we use Qwen2.5-VL-32B-Instruct \citep{Qwen2.5-VL} to produce a caption based on the first frame of each clip. We also utilize the Waymo Perception dataset~\cite{sun2020waymoperception}, extracting 20-second clips from the front camera, along with the HDMap representation, used for training and evaluating our ablation study.

\begin{figure*}[!t]
\centering
    \includegraphics[width=\textwidth]{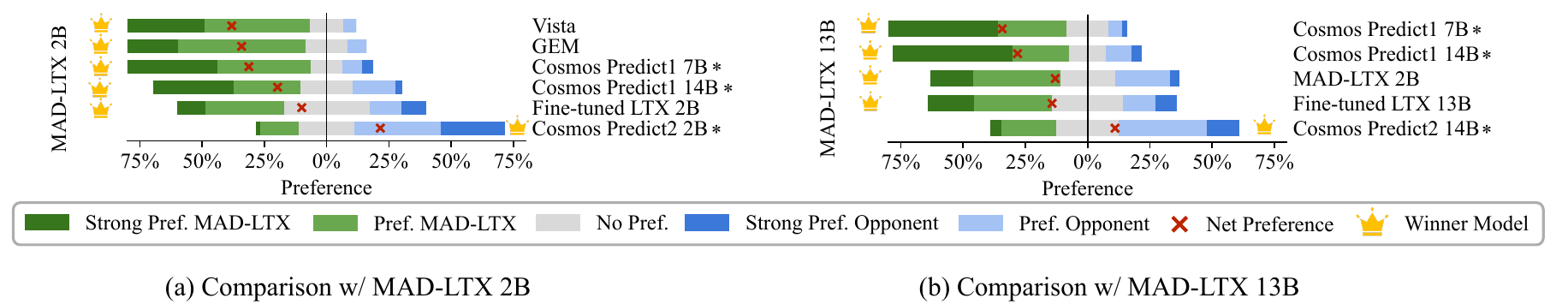}
\caption{
\textbf{Human preference study comparing MAD-LTX against competing models.}
Bars show how often each model is preferred in head-to-head comparisons.
`*' denotes proprietary models which use private datasets and significant computational resources.
}
\label{fig:mad-ltx-pref-combined}
\end{figure*}

\paragraph{Evaluation Metrics:}
We report FID~\cite{heusel2017fid} and FVD~\cite{unterthiner2018fvd} in the Appendix (\autoref{appendix:fid-fvd-numbers}) for completeness. However, as noted by recent works~\cite{ge2024content}, we find these metrics correlate poorly with perceptual quality and motion realism in complex driving scenarios. Our primary evaluation is a human study, where participants were shown videos generated from 100 random scenes from the OpenDV \cite{yang2024opendv} validation set. In a pairwise ``A/B'' test, users were asked to choose which video (from two models, starting from the same initial frame) was superior in: 1) general quality, 2) motion realism, and 3) visual quality. This study collects preference scores across 14 model comparisons. We share a screenshot of the website we developed to collect human preferences in the Supplementary Material (\autoref{appendix:human-study-website}).

\begin{figure}[!t]
\centering
    \includegraphics[width=0.98\columnwidth]{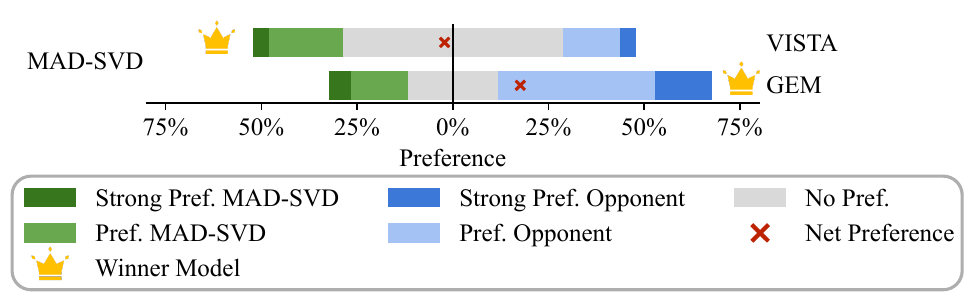}
    \vspace{0.3cm}
\caption{
\textbf{Human preference study comparing MAD-SVD against Vista \cite{gao2024vista} and GEM \cite{hassan2025gem},} which are all based on the save SVD \cite{blattman2023svd} backbone, while MAD-SVD uses only 6\% of Vista and 3\% of GEM's computation budget.
Bars show how often each model is preferred in head-to-head comparisons, which demonstrate that MAD-SVD is able to reach performance of Vista, and get close to GEM's performance only with a fraction of their compute budget.
}
\label{fig:mad-svd-pref}
\end{figure}

\paragraph{Training Details:}
Both the Motion Forecaster ($F_\theta$) and Appearance Synthesizer ($S_\theta$) are initialized from the same base video model (SVD \citep{blattman2023svd} or LTX \citep{hacohen2025ltx}) and fine-tuned using LoRA~\citep{hu2022lora}. We use the AdamW \citep{loshchilov2019adamw} optimizer with a learning rate of $2 \times 10^{-4}$ and a batch size of 32. The motion forecaster $F_\theta$ is trained for 9,000 steps on 139 hours of driving video from OpenDV. The appearance synthesizer $S_\theta$ is subsequently trained for 5,000 steps. All experiments are conducted on 32 GH200 GPUs. Total fine-tuning for our LTX-based model takes 128 GPU hours for the 2B size and 700 GPU hours for 13B size, and 1500 GPU hours for our SVD-based model.
Full training hyperparameters and details are listed in Appendix (\autoref{sec:appendix:training}).

\paragraph{Baselines:}
We apply our methodology to two base models, SVD \cite{blattman2023svd} and LTX \cite{hacohen2025ltx} to create a driving-specific model, ``MAD-SVD'' and ``MAD-LTX'', respectively. We compare our MAD-SVD against existing SVD-based driving models: \textbf{VISTA}~\cite{gao2024vista} and \textbf{GEM}~\cite{hassan2025gem}. We also compare against our state-of-the-art models, the MAD-LTX, with the \textbf{Cosmos}~\cite{nvidia2025cosmosworldfoundationmodel} suite, a state-of-the-art model with two versions: \textit{Predict 1} (7B, 14B), for which a paper and weights are available, and \textit{Predict 2} (2B, 14B), which is a more powerful model with open weights. Since no LTX-based driving models exist for comparison, we establish our own key baseline, Fine-tuned LTX. This model is a LoRA fine-tune of the LTX model on our OpenDV dataset for the same compute budget as our MAD-LTX, trained end-to-end without our two-stage decoupling. This provides a rigorous, `apples-to-apples' comparison to validate our method's efficacy.

\subsection{Proof-of-Concept: MAD-SVD}
\label{sec:exp:SVD-rep}

To first validate our method's efficiency, we implement MAD-SVD by adapting the Stable Video Diffusion (SVD)~\cite{blattman2023svd} model. We fine-tune two separate LoRAs from the SVD backbone: a Motion Forecaster ($F_\theta$) to generate 25 frames of pose video $\mathcal{M}$, and an Appearance Synthesizer ($S_\theta$) to render the final video $\mathcal{X}$.

As shown in \autoref{fig:mad-svd-pref}, we compare MAD-SVD to 
SVD~\cite{blattman2023svd}, VISTA~\cite{gao2024vista} and GEM~\cite{hassan2025gem}, two state-of-the-art driving models also based on SVD. The results are highly encouraging: MAD-SVD achieves competitive generation quality while using a fraction of the resources. Notably, compared to VISTA, our method requires over \textbf{12$\times$ less data} (139 hrs vs.\ 1,700) and \textbf{16$\times$ less compute} (1500 GPU-hrs vs.\ 25,000). This experiment serves as a strong proof-of-concept, demonstrating that our decoupled methodology enables efficient adaptation of general video models into specialized world models, bypassing the need for massive, domain-specific training.

\subsection{Scaling to State-of-the-Art: MAD-LTX}
\label{sec:exp:MAD-LTX}
Having validated our method's efficiency, we now apply it to the large-scale LTX foundation model to create our state-of-the-art driving world model, MAD-LTX. We create and evaluate models at both 2B and 13B scales.

\paragraph{Evaluation of Generation Quality.}
As established in \autoref{sec:exp:setup}, we find that standard metrics like FID and FVD correlate poorly with perceptual quality in this domain (see \autoref{appendix:fid-fvd-numbers}). We therefore rely on our large-scale human preference study as the primary measure of generation quality.

The results of our human study, presented in \autoref{fig:mad-ltx-pref-combined}, lead to three key conclusions:
\begin{enumerate}
    \item \textbf{MAD-LTX outperforms all previous open-source driving models.} At both model scales, MAD-LTX is significantly preferred over SOTA driving models like GEM \citep{hassan2025gem}, VISTA \citep{gao2024vista}, and the proprietary model Cosmos Predict 1 \cite{nvidia2025cosmosworldfoundationmodel}.
    \item \textbf{Our method is superior to naive fine-tuning.} MAD-LTX is strongly preferred over our \textit{Fine-tuned LTX} baseline (29 pref.\ vs.\ 16 for 2B and 33 pref.\ vs.\ 15). This result is critical, as it confirms that our decoupled, two-stage methodology is the primary driver of quality, not just the LTX backbone or the training data.
    \item \textbf{MAD-LTX is competitive with closed-source SOTA.} Our open-source MAD-LTX-13B model achieves generation quality nearly on par with the closed-source Cosmos Predict 2 (14B), while being trained with a fraction of the data and compute, and being significantly faster at inference time.
\end{enumerate}

\paragraph{Evaluation of Open-Loop Motion Planning}

To specifically evaluate the planning capabilities enabled by our Motion Forecaster, we conduct an evaluation designed to isolate the model's open-loop planning ability. We generate 6 unconditional videos (k=6) for each of the 1,500 validation clips. We then use MapAnything~\citep{keetha2025mapanything} to extract the ego-vehicle trajectory from both the ground-truth and generated videos. \autoref{tab:ltx_forecasting} reports the minimum Average Displacement Error (minADE@6) and the Average
Pairwise Distance (APD@6) over a 5-second horizon.
\begin{enumerate} \item \textbf{MAD-LTX achieves superior planning accuracy.} Our model consistently yields the lowest $\text{minADE}_6$, outperforming the Base LTX and our standard Fine-tuned LTX baseline across both scales. 
This performance gain demonstrates that by modeling in the intermediate pose representation space, the Motion Forecaster captures realistic scene interactions and long-term dynamics more effectively than baselines operating directly in video space.

\item \textbf{MAD-LTX is less prone to mode collapse.} We observe that standard fine-tuning significantly degrades diversity of predicted trajectories, with $\text{APD}_6$ lower by 37.8\% comparing to MAD-LTX for the 13B size and causing a 40\% degradation in $\text{minADE}_6$. 
We believe this is caused by the well-known phenomenon of memorization in diffusion models, which is especially likely to occur when large models are fine-tuned on small datasets. While direct video fine-tuning hurts model diversity, MAD-LTX avoids this by predicting motion in an abstract pose space. Because this representation removes visual textures and cues, the model is forced to learn the underlying movement rather than spurious correlations (\eg, always stopping when observing an ambulance, or turning left when seeing a pedestrian wearing pink).
\end{enumerate}

\begin{table}[t]
\centering
\small
\setlength{\tabcolsep}{4pt}
\renewcommand{\arraystretch}{1.1}
\caption{\textbf{Open-loop motion planning evaluation} on OpenDV \citep{yang2024opendv} over a 5s horizon. We evaluate planning accuracy ($\text{minADE}_6$, $\downarrow$ in meters) and trajectory diversity ($\text{APD}_6$, $\uparrow$ in cm). Our MAD-LTX model achieves the best accuracy while maintaining high diversity.}
\label{tab:ltx_forecasting}
\resizebox{\columnwidth}{!}{%
\begin{tabular}{l c c c c}
\toprule
\multirow{2}{*}{Model} & \multicolumn{2}{c}{\textbf{2B} Model} & \multicolumn{2}{c}{\textbf{13B} Model} \\
\cmidrule(lr){2-3} \cmidrule(lr){4-5}
& $\text{minADE}_6 \downarrow$ & $\text{APD}_6 \uparrow$ & $\text{minADE}_6 \downarrow$ & $\text{APD}_6 \uparrow$ \\
\midrule
Base LTX~\citep{hacohen2025ltx} & 5.42 & \textbf{102.96} & 4.14 & \textbf{101.46} \\
Fine-tuned LTX & 5.28 (+2.6\%)  & 68.20 & 5.83(-40\%) & 63.06 \\
MAD-LTX (ours) & \textbf{4.88 (+10\%)}  & 76.21 & \textbf{3.64 (+12\%)} & 101.45 \\
\bottomrule
\end{tabular}}
\end{table}

\subsection{Ablation Studies}
\label{sec:exp:ablations}
We conduct two ablation studies using human preference to validate our primary design choices for the MAD-LTX-2B model.

\begin{table}[t]
\centering
\small
\setlength{\tabcolsep}{4pt}
\renewcommand{\arraystretch}{1.15}
\caption{\textbf{Human preference (\%) for ablations on noise strategy and intermediate representation.} Each value shows the preference for our final MAD-LTX-2B model when compared head-to-head against the variant in the row.}
\label{tab:ablation_uncond}
\resizebox{\columnwidth}{!}{%
\begin{tabular}{@{}lc@{}}
\hline
\multirow{2}{*}{Ablated Model} & Human Pref.\ for MAD-LTX-2B (\%) $\uparrow$ \\
& (Our model is better) \\
\hline
MAD-LTX 2B w/o noise & 62\% \\
MAD-LTX 2B (Panoptic Seg.) & 74\% \\
MAD-LTX 2B (HDMap) & 78\% \\
\hline
\end{tabular}
}

\end{table}

\paragraph{Noise Addition to Appearance Synthesizer.}
To bridge the domain gap between clean ground-truth poses ($\mathcal{M}_{\text{gt}}$) at training and the potentially imperfect predicted poses ($\mathcal{M}_{\text{pred}}$) at inference, we introduce a targeted noising strategy to the $S_\theta$'s conditioning, as detailed in our methodology. \autoref{tab:ablation_uncond} compares our full model against a variant trained ``w/o noise." The strong human preference for our model confirms that this strategy is critical for stabilizing generation and improving the sharpness of the final video.

\paragraph{Intermediate Representation ($\mathcal{M}$).}
A key design choice is the intermediate representation $\mathcal{M}$, as discussed in \autoref{sec:model:motion}. It must be abstract enough for $F_\theta$ to forecast, yet descriptive enough for $S_\theta$ to render. We test two alternatives: 1) \textbf{Panoptic Segmentation} (with tracking) and 2) \textbf{HDMap} (3D bounding boxes with road lanes). The human study in \autoref{tab:ablation_uncond} confirms our pose-based representation is most effective. Our hypothesis is that this strikes an optimal balance: segmentation is pixel-aligned (good for $S_\theta$) but loses 3D orientation (bad for $F_\theta$), while HDMaps are 3D-aware (good for $F_\theta$) but not pixel-aligned (bad for $S_\theta$). Skeleton poses capture the best of both.

\section*{Conclusion}
\label{sec:conclusion}
We introduced MAD, a novel methodology for efficiently adapting general-purpose video foundation models into high-fidelity, controllable driving world models.
By decoupling structured motion learning from appearance synthesis, our framework efficiently converts generalist models into controllable driving world models, bypassing the prohibitive compute and data costs of traditional fine-tuning. 
Our decoupled adaptation is orders of magnitude more efficient than prior work. When scaled, our MAD-LTX model outperforms open-source competitors and achieves generation quality comparable to top-tier proprietary models (Cosmos Predict 2) with a significantly faster inference speed.

Our work presents a highly efficient paradigm for creating specialized, controllable world models from general video generation models. Model weights, code and data collected during our user study will be open sourced.  We hope this work will contribute to reduce the latency between architectural progress in video generation models and their subsequent adoption for driving specialized world models.

\section*{Acknowledgments}
\label{sec:acknowledgments}
 This work was supported as part of the Swiss AI Initiative by a grant from the Swiss National Supercomputing Centre (CSCS) under project ID a144 on Alps. It has also been supported by Hasler Foundation under the Responsible AI program, Swiss National Science Foundation (SNSF) through the project  Grant:10003100, chair VISA DEEP (ANR-20-CHIA-0022), Cluster PostGenAI@Paris (ANR-23-IACL-0007, FRANCE 2030), and Valeo. We also thank the participants who took part in the human evaluation for their time and insights.

{
    \small
    \bibliographystyle{ieeenat_fullname}
    \bibliography{biblio}
}

\clearpage
\setcounter{page}{1}
\maketitlesupplementary

\section{Inference Speed}

\begin{figure}[!t]
  \centering
  \includegraphics[width=1\columnwidth, 
             trim=2cm 2cm 2cm 1.5cm, 
             clip]{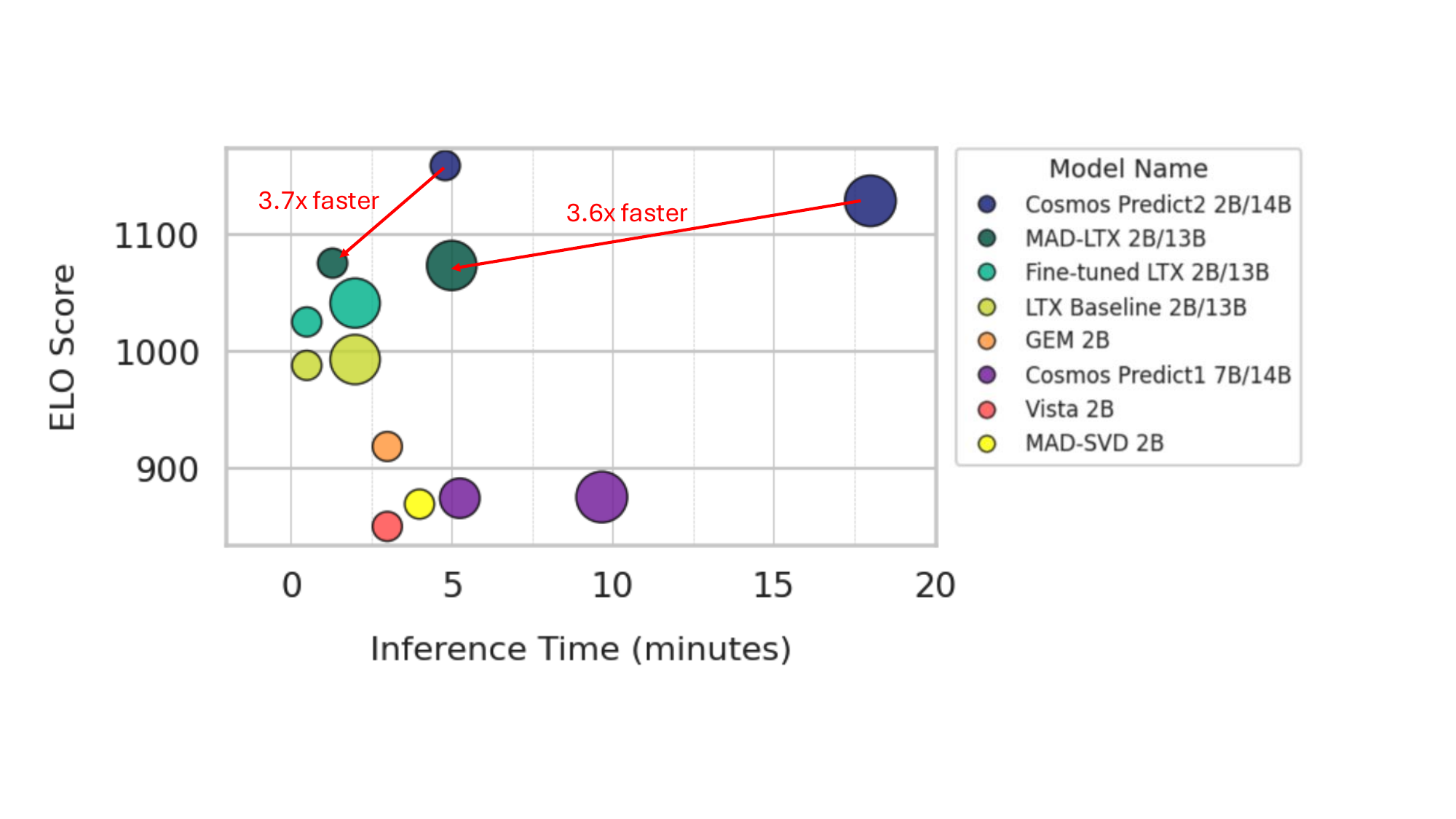}
  \caption{Inference Speed vs. Perceptual Quality. We report the inference time measured on a single GH200 GPU. Performance is quantified using the ELO score computed from our human preference study. \textbf{Our MAD-LTX models achieve up to 3.6× faster inference than competitive models of comparable size and performance}} 
  \label{fig:inf_speed}
\end{figure}

We report the inference latency for 5 seconds video generation in figure \ref{fig:inf_speed} on a single NVIDIA GH200 GPU. We exclude functionalities that require additional model loading (making the inference slower) like upsampling or model offloading for Cosmos baselines. All tests use the default resolution and generated frame count specified by each respective model, for models that generate short segments, multiple rollouts are used to reach the 5-second duration used in our evaluation. 

Direct measurements for Cosmos-Predict2 on the GH200 were not available. We report its A100 inference time as 41 minutes (14B) and 11 minutes (2B). The A100 inference speed is rescaled to the GH200 for Predict2 using a conversion ratio computed from other baselines.

\textbf{Our MAD-LTX models achieve up to 3.6× faster inference than competitive models of comparable size and performance}

\section{Additional Experiments}
\label{appendix:additional_exp}

\subsection{Evaluation of Control Capabilities}
\label{appendix:control_eval}

We evaluate the fidelity of our model's three primary control axes: \textbf{ego trajectory control, object control, and textual control}. Experiments are conducted on a subset of N=800 video clips (5 seconds each) from the Waymo Open Dataset. As shown in \autoref{tab:ltx_control}, we adopt the evaluation protocol from \cite{hassan2025gem} and use our unconditional models (`MAD-LTX-uncond') as a reference baseline to demonstrate the effectiveness of our conditioning signals. The specific evaluation metrics for each modality are defined as follows:

\vspace{1mm} \noindent \textbf{Ego-Motion Control.} We condition the generation on Ground Truth (GT) ego-camera poses from the Waymo Open Dataset. To measure fidelity, we extract the realized ego-trajectory from both the GT video and the generated video using MapAnything \cite{keetha2025mapanything}. We report the \textbf{Average Displacement Error (ADE)} between these two extracted trajectories as the primary metric.

\vspace{1mm} \noindent \textbf{Object Control.} We utilize OpenPifPaf \cite{kreiss2022openpifpaf} to detect and track a single target object in the GT video. We condition the generation on the single object extracted location from the GT video. For evaluation, we run OpenPifPaf \cite{kreiss2022openpifpaf} on the generated video to locate the corresponding object. We report the \textbf{Intersection over Union (IoU)} between the conditioned GT box and the detected box in the generated frame. If the target object fails to appear in the generated video, an IoU of 0 is assigned.

\vspace{1mm} \noindent \textbf{Text Control.} We employ an automated Visual Question Answering (VQA) pipeline to assess text adherence across two distinct categories: \textbf{Action} (dynamics/maneuvers) and \textbf{Object} (entity presence). We use \textbf{Qwen2.5} \cite{qwen2025qwen25technicalreport} to extract semantic elements from the input prompt and reformulate them into a binary verification question tailored to the specific category---for example, \textit{Is the ego car turning left?''} for textual action control, and \textit{Is a cyclist entering the video?''} for textual object control. We then feed the generated video and this question into the Vision-Language Model \textbf{Qwen2.5-VL} \cite{Qwen2.5-VL} to obtain a Yes'' or No'' prediction. We report the \textbf{Success Rate} (percentage of ``Yes'' answers) for each category.

\begin{table}[t]
\centering
\small
\setlength{\tabcolsep}{4pt}
\renewcommand{\arraystretch}{1.1}
\caption{Evaluation of control fidelity. We evaluate each control type over 800 generated samples from Waymo Open Dataset \cite{sun2020waymoperception}. `MAD-LTX-uncond' (our models without any control inputs) serves as  reference baselines to measure the effectiveness of the conditioning. \textbf{MAD-LTX effectively supports all three distinct control modalities}}
\begin{tabular}{@{}l c c c c@{}}
\toprule
\multirow{2}{*}{Model} & \multirow{2}{*}{Ego $\downarrow$} & \multirow{2}{*}{Obj $\uparrow$} & \multicolumn{2}{c}{Text $\uparrow$} \\
\cmidrule(l){4-5} 
 & & & \textbf{Action} & \textbf{Object} \\
\midrule
MAD-LTX-2B-uncond & 5.2 & 0.40 & 38.8\% & 42.1\% \\
MAD-LTX-2B (ours) & \textbf{1.4} & \textbf{0.51} & \textbf{40.8\%} & \textbf{45.6\%} \\
\midrule
MAD-LTX-13B-uncond & 3.4 & 0.45 & 39.2\% & 44.6\% \\
MAD-LTX-13B (ours) & \textbf{1.5} & \textbf{0.55} & \textbf{43.9\%} & \textbf{49.1\%} \\
\bottomrule
\end{tabular}
\label{tab:ltx_control}
\end{table}

\vspace{1mm}
\noindent \textbf{Control Results} 
As detailed in Table~\ref{tab:ltx_control}, we observe consistent quantitative improvements across all metrics and model sizes when comparing the conditioned MAD-LTX models against their unconditional counterparts. The performance gap is most significant in \textbf{Ego-Motion Control amd Object control}. This is expected: while the unconditional model typically forecasts a \textit{plausible} future object/ego trajectory (representing one valid mode of the motion distribution), it is penalized for deviating from the specific Ground Truth realization. The conditioned model, however, effectively utilizes the control signal to lock onto the correct mode. For \textbf{Textual Control}, the improvements are consistently positive but less significant. We attribute this to the unconditional model being trained on static captions (describing the first frame), which yields a high starting score, while VLM noise limits the measured gain.  These results confirm that---consistent across both parameter scales---\textbf{MAD-LTX effectively supports all three distinct control modalities.}

\subsection{Standard video quality metrics (FID/FVD)}
\label{appendix:fid-fvd-numbers}
We report in table \ref{tab:quantitative_results} the Fréchet Inception Distance \cite{heusel2017fid} (FID) and Fréchet Video Distance \cite{unterthiner2018fvd} (FVD) for the Base LTX \cite{hacohen2025ltx} model, the standard fine-tuned baseline (Fine-tuned LTX), and our proposed MAD-LTX across both 2B and 13B parameter scales.

All metrics were computed on the OpenDV \cite{yang2024opendv} test set using N=5,000 generated samples per method. As shown in Table 1, while we report these metrics for completeness, we observe—consistent with recent literature\cite{ge2024content} —that \textbf{FID and FVD scores do not strictly correlate with human perceptual preference in this domain.}

\begin{table}[h]
\centering
\caption{\textbf{Standard video quality metric on the OpenDV test set.} We report FID and FVD metrics computed using $N=5,000$ samples. Consistent with prior work\cite{ge2024content}, \textbf{we observe that these distribution based metrics do not necessarily correlate with human perceptual preference.}}
\label{tab:quantitative_results}
\begin{tabular}{llcc}
\toprule
\textbf{Model Size} & \textbf{Method} & \textbf{FID} $\downarrow$ & \textbf{FVD} $\downarrow$ \\
\midrule
\multirow{3}{*}{2B} & Base LTX & 4.06 & \textbf{64.40} \\
 &  Fine-tuned LTX & \textbf{2.66} & 67.17 \\
 & MAD-LTX (Ours) & 3.72 & 92.79 \\
\midrule
\multirow{3}{*}{13B} & Base LTX & 2.21 & \textbf{56.15} \\
 & Fine-tuned LTX & \textbf{1.94} & 69.56 \\
 & MAD-LTX (Ours) & 2.39 & 59.61 \\
\bottomrule
\end{tabular}
\end{table}

\section{Human Evaluation}
\label{appendix:human-study-website}

\subsection{Protocol}
\begin{figure}[!t]
  \centering
  \includegraphics[width=1\columnwidth, trim=3.9cm 0cm 3.7cm 0, clip]{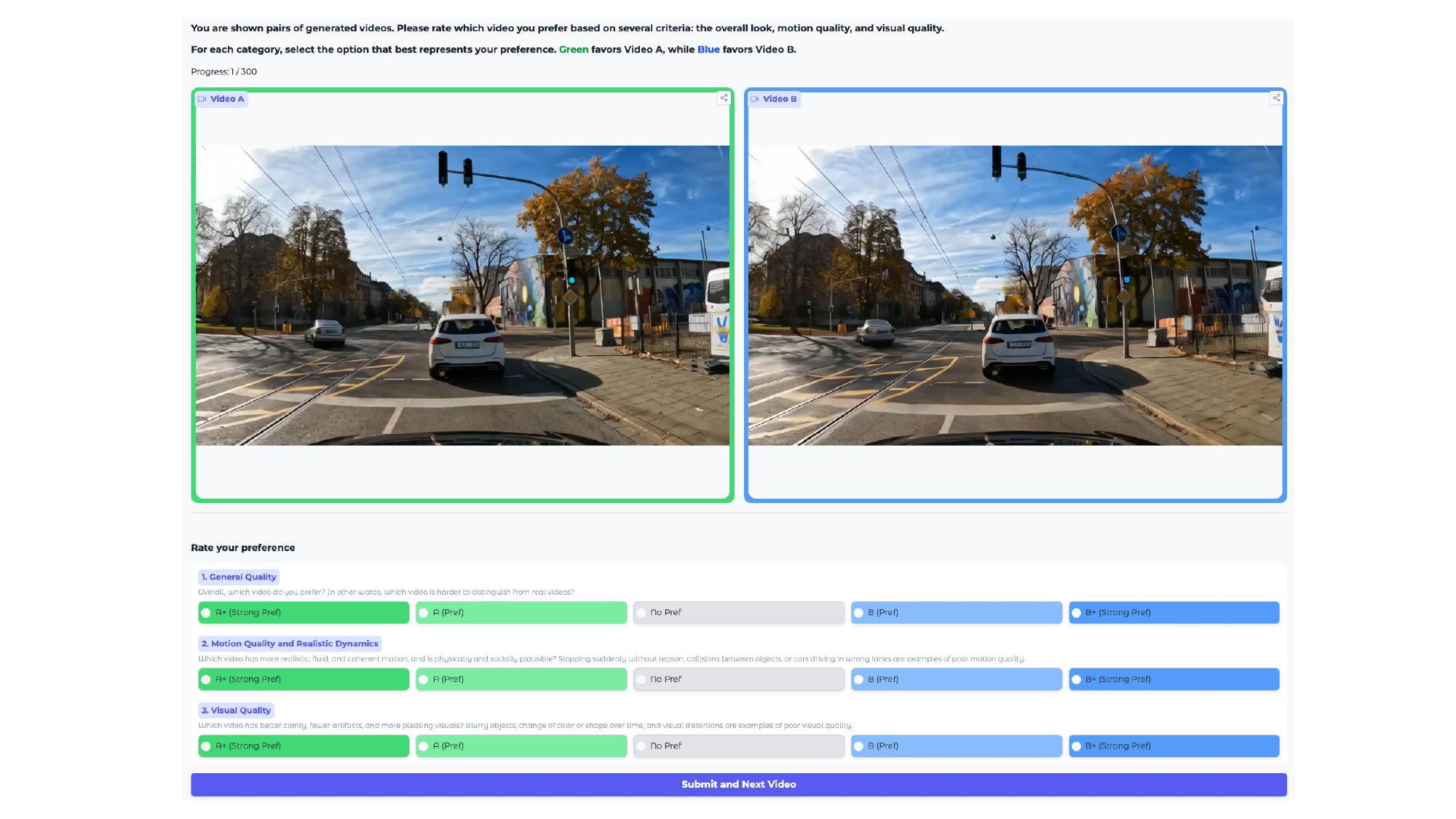}
  \caption{ \textbf{User Study Interface:} We employ a side-by-side pairwise comparison protocol to evaluate generation performance. Participants are asked to rate their preference between two anonymized videos across three specific criteria: General Quality, Motion Quality and Realistic Dynamics, and Visual Quality. For each criterion, annotators select from a 5-point scale ranging from "Strong Preference" or "Preference" for a specific model, to "No Preference."} 
  \label{fig:protocol}
\end{figure}

We conducted a blinded pairwise comparison study hosted on a custom Hugging Face interface. The specific layout of the pairwise comparison interface is illustrated in fugure \ref{fig:protocol}. To ensure high-quality feedback, the participant pool consisted exclusively of domain experts---computer vision researchers specializing in autonomous driving.

For each baseline, we generated $N=100$ video samples (5 seconds duration). In every comparison, annotators were presented with two anonymized videos and asked to rate their preference (Strongly Prefer, Prefer, or No Preference) based on the following specific prompts:

\begin{itemize}
    \item \textbf{General Quality:} \textit{``Overall which video do you prefer? In other words which video is harder to distinguish from real video?''} (Results reported in Figure~5 of the main manuscript).
    
    \item \textbf{Motion Quality and Realistic Dynamics:} \textit{``Which video has more realistic, fluid and coherent motion, and is physically and socially plausible? Stopping suddenly without reason, collisions between objects, or cars driving in wrong direction are example of poor motion quality.''}
    
    \item \textbf{Visual Quality:} \textit{``Which video has better clarity, fewer artifacts, and more pleasing visual? Blurry object, change of color or shape over time, and visual distortions are example of poor visual quality.''}
\end{itemize}

\subsection{Additional results}
In this section, we report the detailed results for \textbf{Motion Quality} and \textbf{Visual Quality}.

\begin{figure}[!t]
  \centering
  \includegraphics[width=1\columnwidth, 
             trim=0cm 0cm 0cm 0cm, 
             clip]{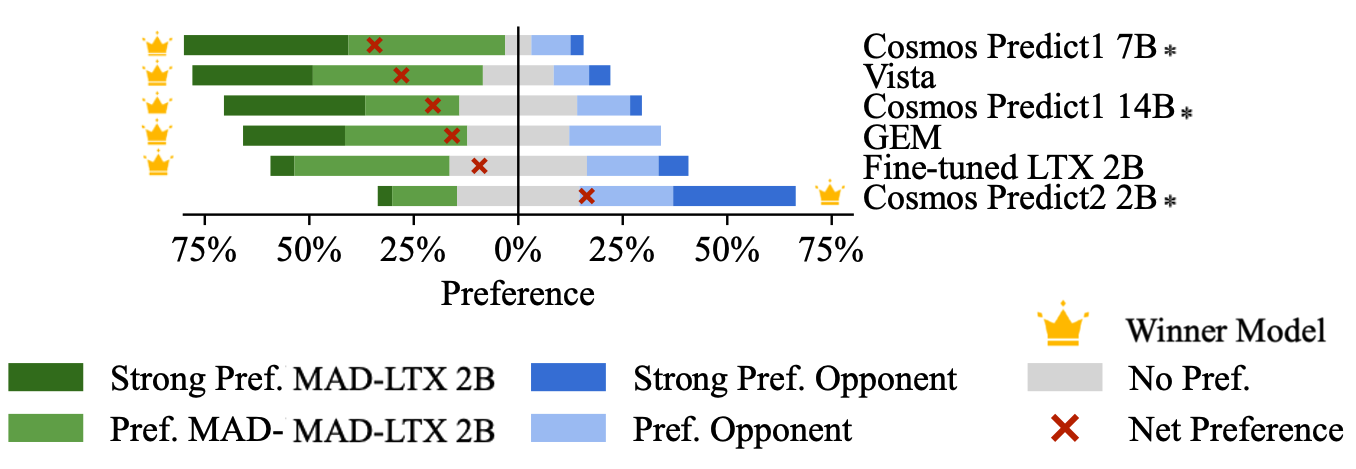}
  \caption{\textbf{Motion Realism Preference (MAD-LTX 2B).} Human preference evaluation specific on motion plausibility (e.g., fluidity, absence of collisions/sudden stops). ‘*’ denotes proprietary mod-
els which use private datasets and significant computational resource \textbf{MAD-LTX-2B outperforms all open-source models, ranking second only to the proprietary Cosmos-Predict2.}}
  \label{fig:2bmotion}
\end{figure}

\begin{figure}[!t]
  \centering
  \includegraphics[width=1\columnwidth, 
             trim=0cm 0cm 0cm 0cm, 
             clip]{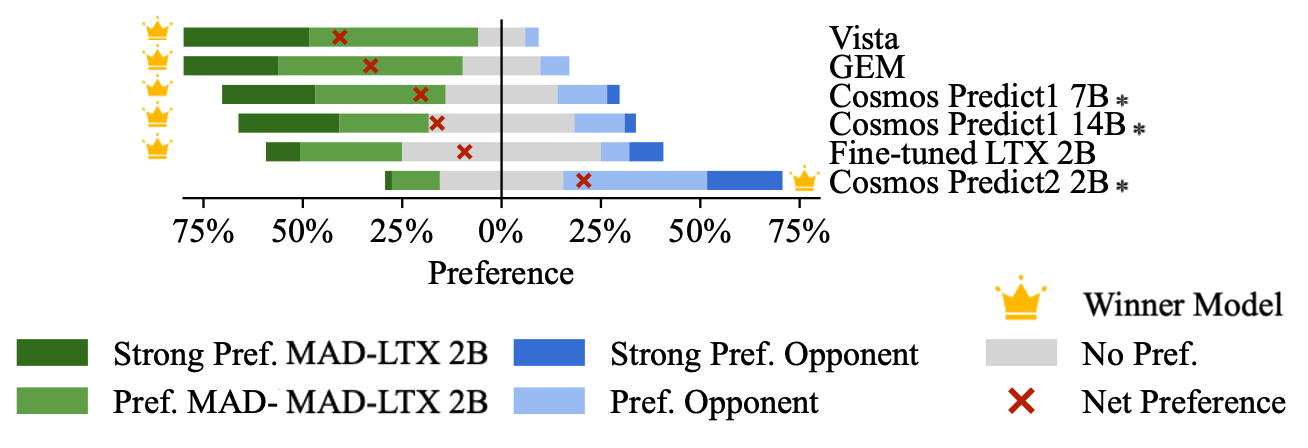}
  \caption{\textbf{Visual Quality Preference (MAD-LTX 2B).} Human preference evaluation specific on visual quality (e.g., temporal consistency, distortions). `*' denotes proprietary models which use private datasets and significant computational resources. \textbf{MAD-LTX-2B outperforms all open-source models, ranking second only to the proprietary Cosmos-Predict2.}}
  \label{fig:2bvisual}
\end{figure}

\begin{figure}[!t]
  \centering
  \includegraphics[width=1\columnwidth, 
             trim=0cm 0cm 0cm 0cm, 
             clip]{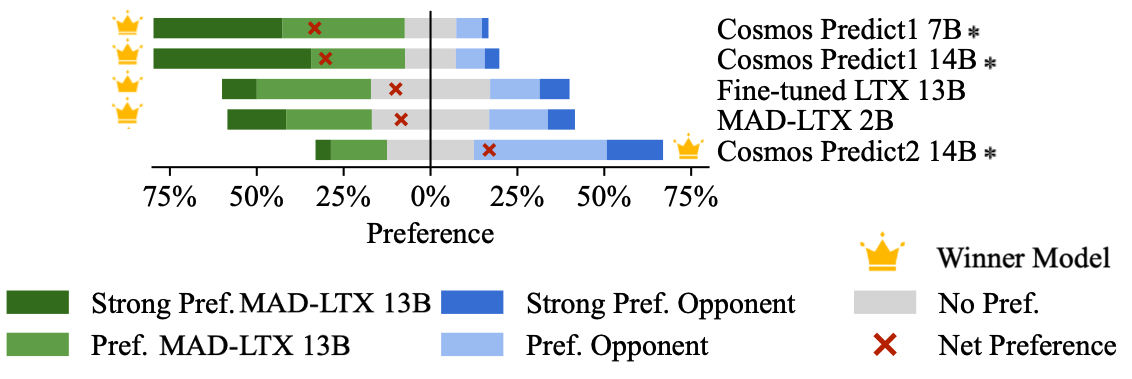}
  \caption{\textbf{Motion Realism Preference (MAD-LTX 13B).} Human preference evaluation specific on motion plausibility (e.g., fluidity, absence of collisions/sudden stops). `*' denotes proprietary models which use private datasets and significant computational resources. \textbf{MAD-LTX-13B outperforms all open-source models, ranking second only to the proprietary Cosmos-Predict2.}}
  \label{fig:13bmotion}
\end{figure}

\begin{figure}[!t]
  \centering
  \includegraphics[width=1\columnwidth, 
             trim=0cm 0cm 0cm 0cm, 
             clip]{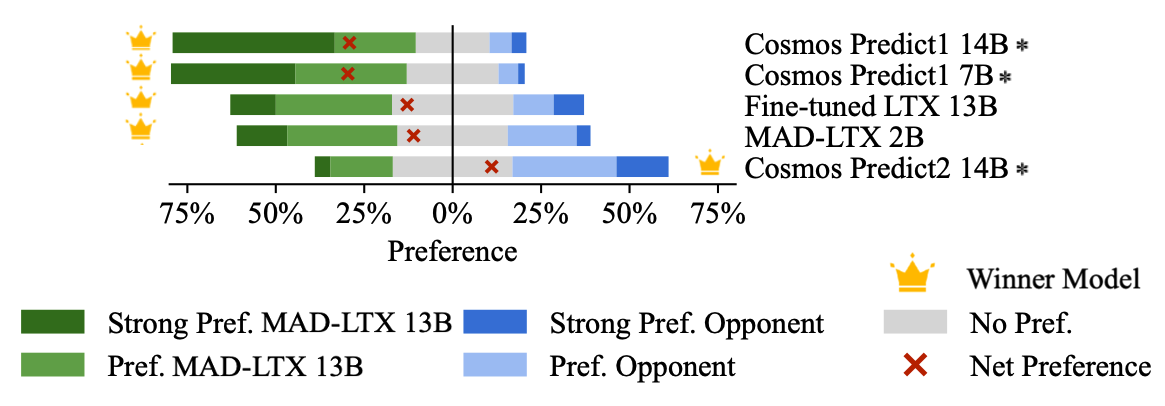}
 \caption{\textbf{Visual Quality Preference (MAD-LTX 13B).} Human preference evaluation specific on visual quality (e.g., temporal consistency, distortions). `*' denotes proprietary models which use private datasets and significant computational resources. \textbf{MAD-LTX-13B outperforms all open-source models, ranking second only to the proprietary Cosmos-Predict2.}}
  \label{fig:13bvisual}
\end{figure}

The detailed pairwise win rates for Motion Quality and Visual Quality are presented in figures~\ref{fig:2bmotion}, \ref{fig:2bvisual}, \ref{fig:13bmotion} and \ref{fig:13bvisual}. We observe that user preferences on these two axes strongly correlate with the General Quality results reported in the main manuscript. Consistent across both the 2B and 13B parameter scales, \textbf{MAD-LTX outperforms all evaluated open-source baselines in terms of both motion realism and visual fidelity.} In alignment with the global preference trends, the proprietary Cosmos-Predict2 is the only model that retains a higher preference rate than our method on both criteria.

\section{Additional Hyperparameters}
\label{sec:appendix:training}
We share a comprehensive list of all the hyper parameters we used for training our Motion Forecaster and our Motion Synthesizer in \autoref{tab:hyperparameters_predictor}. All codes with the corresponding config files will be released for reproducibility.

\begin{table}[htbp]
    \centering
    \caption{LTX Training Hyperparameters}
    \renewcommand{\arraystretch}{1.2} 
    \resizebox{\columnwidth}{!}{
    \begin{tabular}{@{} l l l p{6cm} @{}}
        \toprule
        \textbf{Category} & \textbf{Parameter} & \textbf{Value} & \textbf{Description} \\
        \midrule
        \textbf{Model Strategy} & Model Source & \texttt{LTX\_2B\_0.9.6\_DEV} and \texttt{LTX\_13B\_097\_DEV} & Base 2B parameter model and 13B parameter model. \\
         & Training Mode & \texttt{lora} & Low-Rank Adaptation fine-tuning. \\
         & Precision & \texttt{bf16} & Brain Floating Point 16 mixed precision. \\
        \midrule
        \textbf{LoRA Config} & Rank ($r$) & 512 & High rank for high expressivity. \\
         & Alpha ($\alpha$) & 512 & Scaling factor (1:1 ratio with rank). \\
         & Targets & \texttt{q, k, v, out, ff} & Applies to Attention and Feed-Forward blocks. \\
        \midrule
        \textbf{Optimization} & Learning Rate & $2 \times 10^{-4}$ & Step size for the optimizer. \\
         & Optimizer & \texttt{adamw} & Standard AdamW optimizer. \\
         & Batch Size & 32 & Small batch size due to VRAM constraints. But used 32 GPUs, so an effective batch size of 32. \\
         & Scheduler & \texttt{linear} & Linear learning rate decay. \\
        \bottomrule
    \end{tabular}}
    \label{tab:hyperparameters_predictor}
\end{table}

\section{Limitations and Future Work}
While our intermediate motion representation has proved to be useful, it currently omits contextual driving signalization for precise prediction, such as traffic light states or traffic signs.  Furthermore, we observed that while the overall pseudo-labeling strategy was successful, the generated lane keypoints are inherently noisy, hindering accurate road layout modeling. An ablation showed that replacing these with road segmentation masks could offer a more robust structural representation. For future work, we will focus on incorporating these driving elements and exploring segmentation-based road representations.

Beyond the driving domain, the core MAD principle could be extended to other areas for which pose is a natural representation, such as human-centric video generation, an area where industry currently focuses on extensive training over large datasets, and to ego-view object manipulation for robotics, which directly relates to the world model problematics.

\end{document}